\documentclass[11pt,a4paper]{article}
\usepackage[hyperref]{emnlp2020}
\usepackage{times}
\usepackage{array}
\usepackage{latexsym}

\usepackage{graphicx}
\usepackage{amsmath}
\usepackage{amsthm}
\usepackage{amssymb}
\usepackage{booktabs}
\usepackage{multirow}
\usepackage{algorithmic}
\usepackage[ruled,vlined,linesnumbered]{algorithm2e}
\newcommand{\mrm}[1]{ }

\newcommand{\hl}[1]{\textcolor{red}{#1}}

\usepackage{amsthm,amsmath,amsfonts,bm,xspace}
\usepackage{upgreek}
\usepackage{color}

\def\vs{{\em v.s.}\xspace}
\newcommand{\comment}[1]{}

\def\eqref#1{(\ref{#1})}

\def\1{\bm{1}}

\def\vtheta{{\bm{\theta}}}
\def\va{{\bm{a}}}
\def\vb{{\bm{b}}}

\def\vg{{\bm{g}}}

\def\vq{{\bm{q}}}

\def\vs{{\bm{s}}}

\DeclareMathAlphabet{\mathsfit}{\encodingdefault}{\sfdefault}{m}{sl}
\SetMathAlphabet{\mathsfit}{bold}{\encodingdefault}{\sfdefault}{bx}{n}
\newcommand{\tens}[1]{\bm{\mathsfit{#1}}}

\def\tW{{\tens{W}}}

\usepackage{microtype}

\aclfinalcopy 
\def\aclpaperid{3672} 

\title{Few-Shot Complex Knowledge Base Question Answering \\ via Meta Reinforcement Learning}

\author{Yuncheng Hua$^{\dagger,\S}$, Yuan-Fang Li$^\diamondsuit$, Gholamreza Haffari$^\diamondsuit$, Guilin Qi$^{\dagger,\ddagger}$\thanks{\; Corresponding Author.}\, and Tongtong Wu$^\dagger$\\
$^\dagger$School of Computer Science and Engineering, Southeast University, China\\
$^\diamondsuit$Faculty of Information Technology, Monash University, Australia\\
$^\S$Southeast University-Monash University Joint Research Institute, China\\
$^\ddagger$Key Laboratory of Computer Network and Information Integration, Southeast University\\
\texttt{$^\dagger$\{devinhua, gqi, wutong8023\}@seu.edu.cn}\\
\texttt{$^\diamondsuit$\{yuanfang.li, gholamreza.haffari\}@monash.edu}\\}

\date{}

\begin{document}
\maketitle
\begin{abstract}
%
Complex question-answering (CQA) involves answering complex natural-language questions on a knowledge base (KB).
%
%
However, the conventional neural program induction (NPI) approach exhibits uneven performance when the questions have different types, harboring inherently different characteristics, e.g., difficulty level. 
%
This paper proposes a meta-reinforcement learning approach to program induction in CQA to tackle the potential distributional bias in questions.
Our method quickly and effectively adapts the meta-learned programmer to new questions based on the most similar questions retrieved from the training data.
The meta-learned policy is then used to learn a good programming policy, utilizing the \emph{trial} trajectories and their rewards for similar questions in the support set. 
%
%
Our method achieves state-of-the-art performance on the CQA dataset \cite{saha2018complex} while using only five trial trajectories for the top-5 retrieved questions in each support set, and meta-training on tasks constructed from only 1\% of the training set.
We have released our code at {\small\textsf{\url{https://github.com/DevinJake/MRL-CQA}}}.


\end{abstract}

\section{Introduction}
\label{sec:intro}
%
Knowledge-base question-answering (KBQA) interrogates a knowledge-base (KB)~\cite{yin2016simple,yu2017improved,DBLP:journals/dint/JinLZHLZ19} by interpreting natural-language questions as logical forms (\emph{annotations}), which can be directly executed on the KB to yield answers (\emph{denotations})~\cite{DBLP:conf/acl/PasupatL16}.
KBQA includes \emph{simple questions} that retrieve answers from single-hop triples (``what is Donald Trump's nationality'')~\cite{berant2013semantic,yih2014semantic}, \emph{multi-hop questions} that infer answers over triple chains of at least 2 hops under specific constraints (``who is the president of the European Union 2012'')~\cite{yih2016value,liang2017neural}, and \emph{complex questions} that involve set operations (``how many rivers flow through India and China'')~\cite{saha2019complex}.
In particular, complex question answering (CQA)~\cite{saha2018complex} is a sophisticated KBQA task in which a sequence of discrete \emph{actions}---e.g., set intersection and union, counting, comparison---needs to be executed, and is the subject of this paper.


%
Consider the complex question ``How many rivers flow through India and China?''. We first form a set of entities whose type is {\small\textsf{river}} and {\small\textsf{flow}} in {\small\textsf{China}} from the KB. 
We then form another set for {\small\textsf{rivers}} that {\small\textsf{flow}} through {\small\textsf{India}}.
The answer is then generated by \emph{counting} the entities in the \emph{intersection} of the two sets. 
More concretely, the question is transformed into the action sequence ``{\small{\textsf{Select (China, flow, river), Intersection (India, flow, river), Count}}}'', which is executed on the KB to yield the answer. 
As such, the CQA task results in a massive search space beyond just entities in the KB and includes (lists of) Boolean values and integers.
Multi-hop questions only require the \textsf{join} operator. In contrast, CQA requires various types of additional symbolic reasoning, e.g., logical, comparative, and quantitative reasoning~\cite{shen-etal-2019-multi,ansari2019neural}, where a more diverse array of complex queries is involved~\cite{saha2019complex}. 
The massive search space and complex queries make CQA considerably challenging and more complicated than multi-hop question answering.

Due to the difficulty of collecting annotations, the existing CQA dataset~\cite{saha2018complex} only contains the denotations for each question. 
The literature takes two approaches to deal with the missing annotations. 
The first approach aims to transform learning a CQA model into learning by demonstration, aka imitation learning, where a \emph{pseudo-gold} action sequence is produced for the questions in the training set \cite{guo2018dialog}.
This is done by employing a blind search algorithm, i.e., breadth-first search (BFS), to find a sequence of actions whose execution would yield the correct answer.
This pseudo-gold annotation is then used to train the programmer using teacher forcing, aka behaviour cloning. 
However, BFS inevitably produces a single annotation and is ignorant to many other plausible annotations yielding the correct answer. 
To alleviate this issue, a second approach was proposed based on reinforcement learning (RL) to use the search policy prescribed by the programmer 
\cite{DBLP:journals/ws/Hua20,DBLP:journals/corr/NeelakantanLS15,liang2017neural}. 
Compared to BFS which is a blind search algorithm, the RL-trained programmer can be regarded as an informed search algorithm for target programs. 
Therefore, the RL policy not only addresses the limitation of the 1-to-1 mapping between the questions and annotations, but also produces reasonable programs faster than BFS.

The conventional approach to CQA is to train \emph{one} model to fit the entire training set, and then use it for answering all complex questions at the test time. 
However, such a one-size-fits-all strategy is sub-optimal as the test questions may have diversity due to their inherently different characteristics~\cite{huang2018natural}. 
For instance, in the CQA dataset, the samples could be categorized into seven different types, e.g., those capturing logical/comparative/quantitative reasoning.
The length and complexity of questions in one group are likely to differ from those in other groups.
Therefore, action sequences relevant to different groups may have significant deviations, and it is hard to learn a one-size-fits-all model that could adapt to varied types of questions.
An exception is \cite{DBLP:conf/acl/GuoTDZY19}, which proposes a few-shot learning approach, i.e., S2A, to solve the CQA problem with a retriever and a meta-learner. 
The retriever selects similar instances from the training dataset to form \emph{tasks}, and the meta-learner is trained on these tasks to learn how to quickly adapt to a new task created by the target question of interest at the test time. 
However, Guo et al. \shortcite{DBLP:conf/acl/GuoTDZY19} make use of teacher forcing within the learning by demonstration approach, which suffers from the aforementioned drawbacks.
Also, though S2A is the most similar to ours, the tasks are very different. 
S2A aims to answer \emph{context-dependent} questions, where each question is part of a \emph{multiple-turn} conversation. 
%
On the contrary, we consider the different task where the questions are \emph{single-turn} and have no context.
Thus, a novel challenge arises in retrieving accurate support sets without conversation-based context information.

In this paper, we propose a Meta-RL approach for CQA (MRL-CQA), where the model adapts to the target question by \emph{trials} and the corresponding reward signals on the retrieved instances.
In the meta-learning stage, our approach learns a RL policy across the tasks for both (i) collecting trial trajectories for effective learning, and (ii) learning to adapt programmer by effectively combining the collected trajectories.
%
%
%

The accumulated general knowledge acquired during meta-learning enables the model to generalize over varied tasks instead of fitting the distribution of data points from a single task.
Thus, the tasks generated from tiny (less than 1\%) portion of the training data are sufficient for meta learner to acquire the general knowledge.
Our method achieves state-of-the-art performance on the CQA dataset with overall macro and micro F1 scores of 66.25\% and 77.71\%, respectively.

\section{Meta-RL for Complex Question Answering}
\label{sec:approach}

The problem we study in this paper is transforming a complex natural-language question into a sequence of actions, i.e., a sequence-to-sequence learning task.
By executing the actions, relevant triples are fetched from the KB, from which the answer to the question is induced. 
We tackle this problem with few-shot meta reinforcement learning to decrease the reliance on data annotation and increase the accuracy for different questions.

Let \boldsymbol{$q$} denote the input sequence, including the complex question and the KB artifacts, i.e., entities, relations, and types in KB that are relevant to the problem.
Let \boldsymbol{$\tau$} denotes the output sequence, i.e., an action sequence that the agent generates to answer the question.
Let $R({\boldsymbol{\tau}}|\boldsymbol{q}) \in [0,1]$ denotes the partial reward feedback that tells whether or not the action sequence yields the correct answer.
To simplify the notation, we denote the reward function by $R({\boldsymbol{\tau}})$. 
The training objective is to maximize the expected reward by optimizing the parameter $\vtheta$ of the policy $\pi(\boldsymbol{\tau} | \boldsymbol{q}; \vtheta)$, i.e., improving the accuracy of the policy in answering unseen questions.
For the test, the agent needs to generate an action sequence \boldsymbol{${\tau}^*$} for the input sequence using a search algorithm, e.g., greedy decoding, which is then executed on KB to get the answer.
%


\begin{figure}[t]
\centerline{\includegraphics[width=1.0\columnwidth]{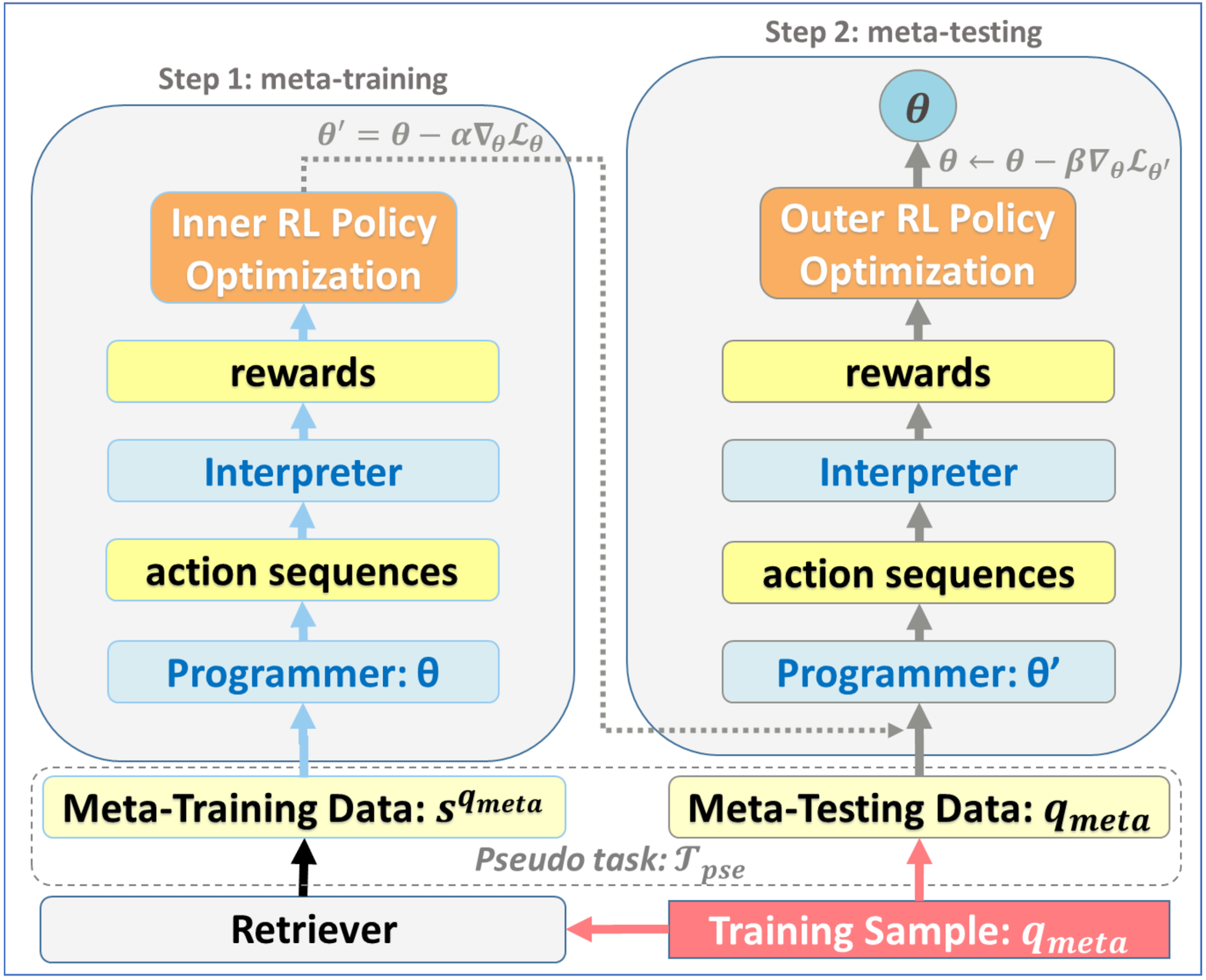}}
\caption{The high-level architecture of our approach.} \label{fig:arch}
\end{figure}

\subsection{Overview of the Framework}
\label{architecture}
Our framework for few-shot learning of CQA is illustrated in Figure~\ref{fig:arch}. 
%
In our framework, we view each new training question as the test sample of a pseudo task, and we aim to learn a specific model devoted to solving the task.
When faced with a question \boldsymbol{$\vq_{meta}$}, we first employ the retriever to find top-N samples \boldsymbol{$\vs^{\vq_{meta}}$} in the training dataset, which are the most similar to \boldsymbol{$\vq_{meta}$}.
We consider \boldsymbol{$\vs^{\vq_{meta}}$} as meta-training data used to learn a particular model, and view the question \boldsymbol{$\vq_{meta}$} as the meta-testing data to evaluate the model.
Therefore, meta-training data \boldsymbol{$\vs^{\vq_{meta}}$} and meta-testing data \boldsymbol{$\vq_{meta}$} form a pseudo task \boldsymbol{$\mathcal{T}_{pse}$}.

In the meta-training step (Step 1 in Figure~\ref{fig:arch}), the action sequences that correspond to \boldsymbol{$s^{q_{meta}}$} will be generated based on the current parameter \boldsymbol{$\theta$} of the \textit{programmer}.
The \textit{interpreter} executes the action sequences and evaluates the generated answers to produce rewards.
The rewards lead to gradient updates that finetune the current model to get a task-specific \textit{programmer} with the parameter of \boldsymbol{${\theta'}$}.
After that, in the meta-testing step (Step 2 in Figure~\ref{fig:arch}), the actions of \boldsymbol{$q_{meta}$} are produced based on \boldsymbol{${\theta'}$} and are evaluated to update \boldsymbol{${\theta}$}.
The training approach is depicted in Algorithm \ref{alg1}.
In both the meta-training and meta-testing steps, REINFORCE~\cite{williams1992simple} is used to optimize the \textit{programmer}.

Similarly, in the inference phase, we consider each test question as a new individual task.
We retrieve top-N data points from the training dataset to form the meta-training data.
Instead of applying the general \textit{programmer} with \boldsymbol{${\theta}$} directly, the meta-training data is used to finetune a specific parameter \boldsymbol{${\theta'}$} that fits the test question and infer the output.

\subsection{Programmer and Interpreter}
\label{programmer_interpreter}
\paragraph{\textbf{Programmer}}
\label{programmer}
Our \textit{programmer} is a sequence-to-sequence (Seq2Seq) model. 
Given the input sequence $\vq$ with tokens $(w_{1},\ldots,w_M)$, the \textit{programmer} produces actions $(a_{1},\ldots,a_{T})$.
The input sequence is the original complex question concatenated with the KB artifacts appear in the query, and the output is the words or tokens. 
The output at each time step is a single token. 

In the \textit{programmer}, the encoder is a Long Short Term Memory (LSTM) network that takes a question of variable length as input and generates an encoder output vector $\boldsymbol{e}_{i}$ at each time step $i$ as: $(\boldsymbol{e}_{i},\boldsymbol{h}_{i}) = LSTM[\phi_{E}(w_{i}),\boldsymbol{h}_{i-1}]$.
Here $\phi_{E}(w_{i})$ is word embedding of token $w_{i}$, and $(\boldsymbol{e}_{i},\boldsymbol{h}_{i})$ is the output and hidden vector of the $i$-th time step. 
The dimension of $\boldsymbol{e}_{i}$ and $\boldsymbol{h}_{i}$ are set as the same.

Our decoder of the \textit{programmer} is another attention-based LSTM model that selects output token $a_{t}$ from the output vocabulary $\mathcal{V}_{output}$.
The decoder generates a hidden vector $\boldsymbol{g}_{t}$ from the previous output token $a_{t-1}$. 
The previous step's hidden vector $\boldsymbol{g}_{t-1}$ is fed to an attention layer to obtain a context vector $\boldsymbol{c}_{t}$ as a weighted sum of the encoded states using the standard attention mechanism.
The current step's $\boldsymbol{g}_{t}$ is generated via $\boldsymbol{g}_{t} = LSTM\{\boldsymbol{g}_{t-1}, [\phi_{D}(a_{t-1}),\boldsymbol{c}_{t}]\}$,
where $\phi_{D}$ is the word embedding of input token $a_{t-1}$. 
The decoder state $\boldsymbol{g}_{t}$ is used to compute the score of the target word $v \in \mathcal{V}_{output}$ as,
\begin{equation}
    \pi(a_t = v|\va_{<t},\vq) = \textrm{softmax}(\tW \cdot \vg_t + \vb)_{v} 
\end{equation}  
where $\tW$ and $\vb$ are trainable parameters, and $\va_{<t}$ denotes all tokens generated before the time step $t$. 
We view all the weights in the \textit{programmer} as the parameter $\vtheta$, thus we have the probability that the \textit{programmer} produces an action sequence \boldsymbol{$\tau$} with tokens $\{a_1,...,a_T\}$ as, %
\begin{equation}
\label{action_prob}
    \pi(\boldsymbol{\tau} | \boldsymbol{q}; \theta) = \prod_{t=1}^{T} \pi(a_t = v|\va_{<t},\vq). 
\end{equation}  
When adapting the policy to a target question, our \textit{programmer} outputs action sequences following the distribution computed by equation \ref{action_prob}.
By treating decoding as a stochastic process, the \textit{programmer} performs random sampling from the probability distribution of action sequences to increase the output sequences' diversity.   

\paragraph{\textbf{Interpreter}}
\label{interpreter}
After the \textit{programmer} generates the entire sequence of actions, the \textit{interpreter} executes the sequence to produce an answer.
It compares the predicted answer with the ground-truth answer and outputs a partial reward.
If the type of the output answer is different from that of the ground-truth answer, the action sequence that generates this answer will be given a reward of 0.
Otherwise, to alleviate the sparse reward problem, the \textit{interpreter} takes the Jaccard score of the output answer set and the ground-truth answer set as the partial reward, and sends it back to update parameters of the \textit{programmer} as the supervision signal.

\subsection{Meta Training and Testing}
\label{training}
We formulate training of the programmer in a RL setting, where an agent interacts with an environment in discrete time steps.
At each time step $t$, the agent produces an action (in our case a word/token) $a_t$ sampled from the policy $\pi(a_t | \va_{<t}, \vq; \vtheta)$, where $\va_{<t}$ denotes the sequence generated by the agent from step 1 to $t-1$, and $\vq$ is the input sequence. 
The policy of the agent is the \textit{programmer}, i.e., LSTM-attention model $M$ with parameter $\vtheta$.
The natural-language question concatenated with the KB artifacts will be fed into the encoder as an input, and a sequence of actions is output from the decoder.
In our work, we regard each action sequence produced by the model as one trajectory. 
The action sequence is therefore executed to yield a generated answer, and the similarity between the output answer with the ground-truth answer is then computed. 
The environment considers the similarity as the reward $R$ corresponding to the trajectory $\boldsymbol{\tau}$ and gives it back to the agent.
In standard RL, the parameter of the policy $\vtheta$ is updated to maximize the expected reward, 
$
\mathbb{E}_{\boldsymbol{\tau} \sim \pi(\boldsymbol{\tau}|{\vq;\vtheta})}[R(\boldsymbol{\tau})].
\label{rl_goal}
$

In our work, answering each question in the training dataset is considered as an individual task, and a model adaptive to a new task is learned from the support set questions.
To make the meta-learned model generalize to all unseen tasks, we sample the tasks following the distribution of tasks in the training dataset.
We first sample a small subset of the questions $Q_{meta}$ from the training dataset and expand the questions into tasks $\mathcal{T}_{meta}$ through retrieving the top-N samples, then extract a batch of tasks $\mathcal{T}'$ from $\mathcal{T}_{meta}$ under the distribution of tasks in $\mathcal{T}_{meta}$ to update parameters. 

To fully use the training dataset and decrease training time, we study how to train a competitive model by using as few training samples as possible.
As we view CQA as a RL problem under few-shot learning conditions, we make use of Meta-RL techniques~\cite{finn2017model} to adapt the programmer to a new task with a few training samples.
Meta-RL aims to meta-learn an agent that can rapidly learn the optimal policy for a new task $\mathcal{T}$. 
It amounts to learn optimized $\vtheta^*$ using $K$ trial trajectories and the rewards for the support set of a new task.

We use the gradient-based meta-learning method to solve the Meta-RL problem such that we can obtain the optimal policy for a given task after performing a few steps of vanilla policy gradient (VPG)~\cite{williams1992simple,sutton2000policy}.
We divide the meta-learning process into two steps to solve a task, namely the meta-training step and the meta-testing step.
Suppose we are trying to solve the pseudo-task $\mathcal{T}_{pse}$, which consists of $N$ meta-training questions $\vs^{\vq_{meta}}$ that are the most similar to the meta-testing sample $q_{meta}$.
The model first generates $K$ trajectories for each question in $\vs^{\vq_{meta}}$ based on $\vtheta$. 
The reward of each trajectory is given by the environment and then subsequently used to compute $\vtheta'$ adapted to task $\mathcal{T}_{pse}$, as 
\begin{equation}
\vtheta' \leftarrow \vtheta+\eta_1 \nabla_{\vtheta} \sum_{\vq \in \vs^{\vq_{meta}}} \mathbb{E}_{\boldsymbol{\tau} \sim \pi(\boldsymbol{\tau}|{\vq;\theta})}[R(\boldsymbol{\tau})]
\label{adapted_theta}
\end{equation}
During meta-testing, another $K'$ trajectories corresponding to question $\vq_{meta}$ are further produced by $\vtheta'$. 
The reward of $K'$ trajectories are considered as the evaluation of the adapted policy $\vtheta'$ for the given task $\mathcal{T}_{pse}$; thus we have the objective, 
\begin{equation}
J(\vtheta')\overset{\textrm{def}}{=} \mathbb{E}_{\boldsymbol{\tau}' \sim \pi({\boldsymbol{\tau}}'|\vq_{meta};\vtheta')}[R({\boldsymbol{\tau}}')]
\label{meta_test_loss}
\end{equation}
The parameter of the generic policy $\vtheta$ are then trained by maximising the objective $J(\vtheta')$, 
\begin{equation}
\vtheta\leftarrow\vtheta + \eta_2 \nabla_{\vtheta}J(\vtheta')
\label{updated_theta}
\end{equation}
In each VPG step, since we have $N$ samples in $\vs^{\vq_{meta}}$, we use $N$ policy gradient adaptation steps to update $\vtheta'$.
Meanwhile, we use one policy gradient step to optimize $\theta$ based on the evaluation of $\vtheta'$.
Monte Carlo integration is used as the approximation strategy in VPG~\cite{guu2017language}.
We summarise the meta-learning approach in Alg.\ref{alg1}. 

When making inferences, for each question $\vq_{test}$, the retriever creates a pseudo-task, similar to the meta-learning process. 
The top-$N$ similar questions to $\vq_{test}$ form the support set $\vs^{\vq_{test}}$, and are used to obtain the adapted model $\vtheta^{*'}$, starting from the meta learned policy $\vtheta^*$.
The adapted model is then used to generate the program and compute the target question's final answer.

\begin{algorithm}[htb]
\caption{Meta-RL (training time)}
\label{alg1}
\DontPrintSemicolon
\KwIn{Dataset $Q_{train}$, step size $\eta_1$, $\eta_2$}    
\KwOut{Meta-learned policy $\vtheta^*$}
Randomly initialize $\vtheta$\;
Randomly sample $Q_{meta} \sim Q_{train}$\; 
Expand $Q_{meta} \rightarrow \mathcal{T}_{meta}$\;
\While{not done}{
    Sample a batch of tasks $\mathcal{T}' \sim \mathcal{T}_{meta}$\;
    \For{$\mathcal{T}_{pse} \in \mathcal{T}'$}
    {$\mathcal{L}\leftarrow0$\;
    \For{each question $\vq \in \vs^{\vq_{meta}}$}
        {
        Sample $K$ trajectories: $\boldsymbol{\tau}_k \sim \pi(\boldsymbol{\tau}|\vq;\vtheta)$ \;
         $\mathcal{L}\leftarrow\mathcal{L}+\frac{1}{K}{\sum_{k=1}^K R(\boldsymbol{\tau_k})log p_\vtheta(\boldsymbol{\tau_k})}$\;
        }
        $\vtheta'\leftarrow \vtheta+\eta_1 \nabla_{\vtheta}\mathcal{L}$\; 
        Sample $K'$ trajectories: $\boldsymbol{\tau}_{k'} \sim \pi(\boldsymbol{\tau}|\vq_{meta};\vtheta')$ \;
        $J_{\vq_{meta}}(\vtheta') \leftarrow  \frac{1}{K'}\sum_{k'=1}^{K'}R({\boldsymbol{\tau}_{k'}})log p_{\vtheta'}(\boldsymbol{\tau}_{k'})$\;
    }
    $\vtheta\leftarrow\vtheta+\eta_2\nabla_{\vtheta}\sum_{\mathcal{T}_{pse} \in \mathcal{T}'} J_{\vq_{meta}}(\vtheta')$\;
}
\textbf{Return}  The meta-learned policy $\vtheta^* \leftarrow \vtheta$ 
\end{algorithm}


\subsection{Question Retriever}
\label{retriever}
We propose an unsupervised retriever that finds, from the training dataset, relevant support samples for the tasks in both the training and test phases.
%
We propose a relevance function that measures the similarity between two questions in two aspects: (1) the number of KB artifacts (i.e., entities, relations, and types) in the questions and (2) question semantic similarity.

If the two questions have the same number of KB artifacts, the structure of their corresponding action sequences are more likely to be resembled. 
We calculate the similarity in terms of the number of entities of two questions $\vq_1$ and $\vq_2$ by $sim_{e}(\vq_1, \vq_2) = 1-\frac{|q_{e}(\vq_1)-q_{e}(\vq_2)|}{\max(q_{e}(\vq_1),q_{e}(\vq_2))}$. 
The function $q_{e}(\vq)$ counts the number of entities in the question.
Similarly, we compute the similarities in terms of relations and types in the same way with $sim_{r}(\vq_1, \vq_2)$ and $sim_{t}(\vq_1, \vq_2)$ respectively. 
The KB artifact similarity $sim_{a}(\vq_1, \vq_2)$ is computed by the product of the above three similarities.

For two questions, the more common words they have, the more semantically similar they are. 
Based on this intuition, we propose a semantic similarity function based on Jaccard similarity in an unsupervised way. 
Suppose there is a set of $i$ words $\{w_1^1,...,w_1^i\}$ in $\vq_1$ and $j$ words $\{w_2^1,...,w_2^j\}$ in $\vq_2$, and word similarity $sim(w^i, w^j)$ is calculating using the Cosine similarity.



For each word in $\vq_1$, we first collect the word pairs from the words in $\vq_2$, whose highest similarity exceeds a pre-defined threshold value. 
We denote with $sem_{int}(\vq_1,\vq_2)$ the sum of similarity values of the word pairs:
\begin{equation}
sem_{int}(\vq_1, \vq_2) = \sum_{m=1}^{i}(\max_{n=1}^j(sim(w_1^m, w_2^n)))
\label{sem_inter}
\end{equation}

After removing this set of highly similar words from the two questions, we denote the remaining tokens as $\{w_1^{remain}\}$ and $\{w_2^{remain}\}$, which represent the different parts of the two questions.
We sum up the embeddings of the words in $\{w_1^{remain}\}$ as $\mathbf{w}_1^{remain}$, and compute $\mathbf{w}_2^{remain}$ in the same way.
The function $sem_{diff}(\vq_1,\vq_2)$ measures how different $\vq_1$ and $\vq_2$ are: 
\begin{equation}
\begin{aligned}
sem_{diff}(\vq_1, \vq_2) = \max(|\{w_1^{remain}\}|,|\{w_2^{remain}\}|)\\
* (1-sim(\mathbf{w}_1^{remain}, \mathbf{w}_2^{remain})), 
\end{aligned}
\label{sem_diff}
\end{equation}
where $|\{w\}|$ returns the cardinality of the set $\{w\}$.

We define the semantic similarity between $\vq_1$ and $\vq_2$ as: $sim_{s}(\vq_1, \vq_2) = \frac{sem_{int}(\vq_1, \vq_2)}{sem_{int}(\vq_1, \vq_2)+sem_{diff}(\vq_1, \vq_2)}$, and therefore calculate the similarity between $\vq_1$ and $\vq_2$ with  $sim_{a}(\vq_1, \vq_2) * sim_{s}(\vq_1, \vq_2)$.



\section{Experiments}
\label{sec:experiments}
In this section, we present the empirical evaluation of our MRL-CQA framework. 

\paragraph{Dataset.}
We evaluated our model on the large-scale CQA (Complex Question Answering) dataset~\cite{saha2018complex}. 
Generated from the Wikidata KB~\cite{Vrandecic2014WikidataAF}, CQA contains 944K/100K/156K QA pairs for training, validation, and testing, respectively. 
In the CQA dataset, each QA pair consists of a complex, natural-language question and the corresponding ground-truth answer (i.e., denotation). 
We note that annotations, i.e., gold action sequences related to the questions, are not given in the CQA dataset. 
The CQA questions are organized into seven categories of different characteristics, as shown in the Table~\ref{tab2}. 
Some categories have entities as answers (e.g., ``Simple Question''), while others have (lists of) numbers (e.g., ``Quantitative (Count)'') or Booleans (e.g., ``Verification (Boolean)'') as answers. 
The size of different categories in CQA is uneven.
The number of instances in each category in the training set is 462K, 93K, 99K, 43K, 41K, 122K, and 42K for Simple Question, Logical Reasoning, Quantitative Reasoning, Verification (Boolean), Comparative Reasoning, Quantitative (Count), and Comparative (Count), respectively.

Based on the length of the induced programs and performance of the best models, we further organized the seven categories into two groups: \emph{easy}---the first four categories, and \emph{hard}---the last three types, in Table~\ref{tab2}. 
We used the same evaluation metrics employed in the original paper~\cite{saha2018complex}, the F1 measure, to evaluate models.

\paragraph{Training Configuration.}
In the CQA dataset, since the annotated action sequence are not provided, we randomly sampled 1\% of the training set (approx.\ 10K out of 944K training samples) and followed~\cite{DBLP:conf/acl/GuoTDZY19} to annotate them with pseudo-gold action sequences by using a BFS algorithm.
We denoted the 1\% questions and relevant pseudo-gold action sequences as $Q_{pre}$.
The $Q_{pre}$ was used to train the LSTM-based programmer, which was further optimized through the Policy Gradient (PG) algorithm~\cite{williams1992simple,sutton2000policy} with another 1\% unannotated questions from the training set.
We denoted this model by \textbf{PG}, which is also a model variant proposed in~\cite{DBLP:journals/ws/Hua20}. 
We trained the meta learner on another 2K training samples ($Q_{meta}$ in Alg.\ref{alg1}), representing only approx.\ 0.2\% of the training set.
This meta learner is our full model: \textbf{MRL-CQA}.

In our work, we chose the attention-based LSTM model instead of the Transformer~\cite{vaswani2017attention} to design the \textit{programmer}.
We set the sizes of embedding and hidden units in our LSTM model as 50 and 128 respectively, thus the maximum number of the parameters in our model is about 1.2M. 
However, the base model of the Transformer (12 layers, 12 heads, and 768 hidden units) has 110M parameters~\cite{Wolf2019HuggingFacesTS}, which are much more than those of our model.
Given the small size of the training samples and the weak supervision signal (reward in our work), it is harder to train the model with more parameters.
Therefore we chose LSTM rather than the Transformer.

We implemented our model in PyTorch and employed the Reptile meta-learning algorithm to optimize the meta-learned policy~\cite{nichol2018reptile}. 
The weights of the model and the word embeddings were randomly initialized and further updated within the process of training.
In meta-learning, we set $\eta_1 = 1e-4$ (Equation~\ref{adapted_theta}) and $\eta_2 = 0.1$ (Equation~\ref{updated_theta}). 
We set $N = 5$ and threshold value at $0.85$ when forming the support set. 
For each question, we generate five action sequences to output the answers.
The Adam optimizer is used in RL to maximizes the expected reward. 



Among the baseline models, we ran the open-source code of KVmem~\cite{saha2018complex} and CIPITR~\cite{saha2019complex} to train the model.
As the code of NSM~\cite{liang2017neural} has not been made available, we re-implemented it and incorporated our programmer to predict programs, and employed the reinforcement learning settings in NSM to optimize the programmer.
When inferring the testing samples, we used the top beam, i.e., the predicted program with the highest probability in the beam to yield the answers.
We presented the best result we got to compare the baseline models.


\begin{table*}[htb]
\centering
\caption{Performance comparison (measured in F1) of the seven methods on the CQA test set. For each category, best result is \textbf{bolded} and second-best result is \underline{underlined}.
}\label{tab2}
\centering
\begin{tabular}%
{
l*{6}{c}
}
\toprule
{\textbf{Question category}} & {\textbf{KVmem}} & {\textbf{NSM}} & {\textbf{CIP-All}} & {\textbf{CIP-Sep}} & \textbf{PG} & \textbf{MRL-CQA}\\ 
\midrule
Simple Question         & 41.40\%   & \underline{88.83\%}   & 41.62\%   & \textbf{94.89\%}  & 85.20\%               & 88.37\% \\
Logical Reasoning       & 37.56\%   & 80.21\%               & 21.31\%   & \textbf{85.33\%}  & 78.23\%               & \underline{80.27\%} \\
Quantitative Reasoning  & 0.89\%    & 36.68\%               & 5.65\%    & 33.27\%           & \underline{44.22\%}   & \textbf{45.06\%} \\
Verification (Boolean)  & 27.28\%   & 58.06\%               & 30.86\%   & 61.39\%           & \underline{84.42\%}   & \textbf{85.62\%} \\
\midrule
Comparative Reasoning   & 1.63\%    & \underline{59.45\%}   & 1.67\%    & 9.60\%            & 59.43\%               & \textbf{62.09\%} \\
Quantitative (Count)    & 17.80\%   & 58.14\%               & 37.23\%   & 48.40\%           & \underline{61.80\%}   & \textbf{62.00\%} \\
Comparative (Count)     & 9.60\%    & 32.50\%               & 0.36\%    & 0.99\%            & \underline{38.53\%}   & \textbf{40.33\%} \\
\midrule
Overall macro F1        & 19.45\%   & 59.12\%               & 19.82\%   & 47.70\%           & \underline{64.55\%}   & \textbf{66.25\%} \\
Overall micro F1        & 31.18\%   & 74.68\%               & 31.52\%   & 73.31\%           & \underline{75.40\%}   & \textbf{77.71\%} \\
\bottomrule
\end{tabular}
\end{table*}

\subsection{Model Comparisons}
We evaluated our model, MRL-CQA, against three baseline methods on the CQA dataset: KVmem, NSM, and CIPITR.
It must be pointed out that CIPITR separately trained \emph{one single model for each of the seven question categories}. 
We denote the model learned in this way as \textbf{CIP-Sep}.
CIPITR also trained \emph{one single model over all categories of training instances} and used this single model to answer all questions. 
We denote this single model as \textbf{CIP-All}. 
We separately present the performance of these two variations of CIPITR in Table~\ref{tab2}.
On the contrary, we tuned MRL-CQA on all categories of questions with one set of model parameters. 







Table~\ref{tab2} summarizes the performance in F1 of the six models on the test set of CQA, organised into seven question categories. 
We note that the first four categories (first four rows in Table~\ref{tab2}) are relatively simple, and the last three (middle three rows) are more challenging. 
We also report the overall macro and micro F1 values (last two rows). 

As can be seen, our MRL-CQA model achieves the overall best macro and micro F1 values, achieving state-of-the-art results of 66.25\% and 77.71\%, respectively. 
MRL-CQA also achieves the best or second-best performance in six out of the seven categories.
Of the three hardest categories (the last three types in Table~\ref{tab2}), MRL-CQA delivers the best performance in all three types.
This validates the effectiveness of our meta-learning-based approach in effectively learning task-specific knowledge.
Note that the two categories that MRL-CQA performs the best, Comparative Reasoning and Comparative (Count), both account for less than 5\% of the training set, which further demonstrates our model's excellent adaptability.

Also, our RL-based programmer PG achieves second-best result in overall macro and micro F1, with about 2\% difference below MRL-CQA. 
Moreover, PG achieves second-best in four categories. 
Such strong performance indicates the effectiveness of our CQA framework.

Besides the above main result, several important observations can be made from Table~\ref{tab2}.


1. CIP-Sep got the best result in two categories, i.e., ``Simple Question'' and ``Logical Reasoning''.
However, it performed poorly for the three hard categories.
Consequently, the overall macro F1 value of CIP-Sep is substantially lower than both PG and MRL-CQA.
Note that CIP-Sep trained a different model separately for each of the seven question categories. 
The results reported for each category were obtained from the models tuned specifically for each category~\cite{saha2019complex}, which necessitated a classifier to be trained first to recognize the question categories.
Thus, CIP-Sep needs to re-train the models to adapt to new/changed categories, which impedes it from generalizing to unseen instances. 
However, we tuned our models on all questions with one set of model parameters, disregarding the question category information. 



2. As presented in Table~\ref{tab2}, CIP-All, the model that trained over all types of the questions, performed much worse in all the categories than CIP-Sep.
A possible reason for CIP-All's significant performance degradation is that it is hard for such a one-size-fits-all model to find the weights that fit the training data when the examples vary widely.
Besides, the imbalanced classes of questions also deteriorate the performance of the model. 
Different from CIPITR, MRL-CQA is designed to adapt appropriately to various categories of questions with one model, thus only needs to be trained once.




3. Our programmer and carefully-defined primitive actions presented in this work were used in our re-implementation of NSM. 
In the hard categories, by comparing the F1 scores of PG and MRL-CQA, it could be observed that NSM performed competitively.
Furthermore, NSM performed the second best in ``Simple Question" and ``Comparative Reasoning" categories. 
This helps to validate the effectiveness of our proposed techniques. 
However, NSM is worse than MRL-CQA in six out of the seven categories.
This verifies the adaptability of our model, which can quickly adapt to new tasks by employing the learned task-specific knowledge.

Note that our model was trained only on 1\% of the training set, whereas the baseline models use the entire training set.
Besides, our method trains one model to solve all questions, while CIP-Sep trains seven models, one for each category of problems. 
Thus our model is compared with seven individually trained models in CIPITR but still achieves the best overall performance, demonstrating the effectiveness of our technique.

\begin{table}[htb]
\caption{Ablation study on the test set on macro F1 score change with different sizes of training samples.}\label{tab3}
\centering
\begin{tabular}{lr}
\toprule
{\textbf{Feature}} & {\textbf{Overall macro F1}}\\ 
\midrule
PG & 75.40\% \\
\midrule
MRL-CQA \\
\quad 500-training   & +0.01\% \\
\quad 1,000-training & +0.58\% \\ 
\quad 2,000-training & +2.31\% \\
\bottomrule
\end{tabular}
\end{table}

\subsection{Model Analysis}
We conduct an ablation experiment to study the effect of meta-learning. 
We also study the effect of smaller training samples by comparing MRL-CQA's performance trained on 500 and 1K samples, against 2K used in the full model. 

Table~\ref{tab3} summarizes the ablation study result. 
Trained on 500 samples only, MRL-CQA slightly improves performance by 0.01 percentage points compared to PG. 
When training sample increases to 1K, MRL-CQA outperforms PG by 0.58 percentage points. 
The full MRL-CQA model, trained on 2K samples, achieves a performance improvement over PG of 2.31 percentage points. 
These results demonstrate the ability to design a specific model for answering each question precisely, which is afforded by meta-learning. 

\subsection{Case Study}
We provide a case study of different types of questions that MRL-CQA could answer, but our RL-based model, aka PG, could not solve. 
The comparison is given in Table~\ref{tab:case_study}, which lists the action sequences and the corresponding results these two models predicted when answering the same questions. 
We highlight the different parts of the action sequences that the two models generated.

%
For example, when answering the Logical Reasoning question in Table~\ref{tab:case_study}, PG was confused about what relations should be used to form feasible actions.
It could be seen that PG failed to distinguish the two different relations for the two actions and thus produced a wrong answer.

Similarly, when answering the Verification question in Table~\ref{tab:case_study}, PG also yielded an infeasible action sequence.
After forming a set of {\small\textsf{political
territories}} that {\small\textsf{Hermine Mospointner}} is {\small\textsf{a citizen of}}, the {\small\textsf{bool}} action should be used to judge whether {\small\textsf{Valdeobispo}} and {\small\textsf{Austria}} are within the set.
It can be seen that PG missed one action: \emph{Bool (Austria)}.

The different optimization goals lead to the different results of the two models.
PG, as a typical one-size-fits-all model, aims to estimate the globally optimal parameters by fitting itself to the training samples.
Such a model extracts the information from the training data to update model parameters, applies the parameters to the new samples without modification thereafter. 
Therefore, when facing a wide variety of questions, it is hard for the model to find a set of parameters that fits all samples.
Under the circumstances, like what is presented in Table~\ref{tab:case_study}, such a one-size-fits-all model could not handle the questions well.

However, our MRL-CQA model aims to learn general knowledge across tasks and fix the knowledge into the initial parameters.
We thus learn a model that can subsequently adapt the initial parameters to each new given question and specialize the adapted parameters to the particular domain of the new questions.
Therefore, with the help of the adapted parameters, MRL-CQA can answer each new question more precisely than PG.

\begin{table*}[htb]
\centering\small
\caption{A comparison of the action sequences and results that PG (the second column) and MRL-CQA (the third column) yield when answering the same questions.}
\label{tab:case_study}
\begin{tabular}{|p{5cm}|p{5cm}|p{5cm}|}
\hline\hline
\multicolumn{3}{c}{An Example of Logical Reasoning Question} \\ \hline\hline
Question Information & PG & MRL-CQA \\ \hline
\multirow[t]{4}{=}{\textbf{Question:} Which occupations are the professions of Sergio Piacentini \textbf{or} were a position held by Antoinette Sandbach?} & \textbf{Action sequence:} & \textbf{Action sequence:} \\
& \emph{Select (Sergio Piacentini, \hl{position\textunderscore held}, occupation)} & \emph{Select (Sergio Piacentini, \hl{occupation\textunderscore of}, occupation)} \\
& \emph{Union (Antoinette Sandbach, position\textunderscore held, occupation)} & \emph{Union (Antoinette Sandbach, position\textunderscore held, occupation)} \\ \hline
\textbf{Ground-truth answer:} & \textbf{Execution result:} & \textbf{Execution result:} \\
Member of the National Assembly for Wales, & Member of the National Assembly for Wales & Member of the National Assembly for Wales, \\ 
association football manager, & & association football manager, \\ 
association football player & & association football player \\ \hline
\hline
\multicolumn{3}{c}{An Example of Verification (Boolean) Question} \\ \hline\hline
Question Information & PG & MRL-CQA \\ \hline
\multirow[t]{4}{=}{\textbf{Question:} Is Hermine Mospointner a civilian of Valdeobispo \textbf{and} Austria?} & \textbf{Action sequence:} &  \textbf{Action sequence:} \\
& \emph{Select (Hermine Mospointner, country\textunderscore of\textunderscore citizenship, political territory)} & \emph{Select (Hermine Mospointner, country\textunderscore of\textunderscore citizenship, political territory)} \\
& \emph{Bool (Valdeobispo)} & \emph{Bool (Valdeobispo)} \\
& & \emph{\hl{Bool (Austria)}} \\ \hline
\textbf{Ground-truth answer:} & \textbf{Execution result:} & \textbf{Execution result:} \\
False and True & False & False and True \\ \hline
\end{tabular}
\end{table*}

\section{Related Work}
\label{sec:related}

{\bf Imitation Learning.}
Imitation Learning aims to learn the policy based on the expert's demonstration by supervised learning. 
Saha et al. \shortcite{saha2018complex} propose a CQA model that combines Hierarchical Recurrent Encoder-Decoder (HRED) with a Key-Value memory (KVmem) network and predicts the answer by attending on the stored memory.
%
Guo et al. \shortcite{guo2018dialog} present a Dialog-to-Action (D2A) approach to answer complex questions by learning from the annotated programs.
D2A employs a deterministic BFS procedure to label questions with pseudo-gold actions and trains an encoder-decoder model to generate programs by managing dialog memory.
%
Multi-task Semantic Parsing (MaSP)~\cite{shen-etal-2019-multi} jointly optimizes two modules to solve the CQA task, i.e., entity linker and semantic parser, relying on annotations to demonstrate the desired behaviors. 
Different from the above approaches, our model performs better while removing the need for annotations.

{\bf Neural Program Induction (NPI).}
NPI is a paradigm for mapping questions into executable programs by employing neural models.
Neural-Symbolic Machines (NSM)~\cite{liang2017neural} is proposed to answer the multi-hop questions.
NSM annotates the questions and then anchors the model to the high-reward programs by assigning them with a deterministic probability.
Neural-Symbolic Complex Question Answering (NS-CQA) model~\cite{DBLP:journals/ws/Hua20} augments the NPI approach with a memory buffer to alleviate the sparse reward and data inefficiency problems appear in the CQA task.
Complex Imperative Program Induction from Terminal Rewards (CIPITR)~\cite{saha2019complex} relies on auxiliary awards, KB schema, and inferred answer types for training an NPI model to solve the CQA task.
However, CIPITR separately trains one model for each category of questions with a different difficulty level. 
Compared with the NPI models, our model can flexibly adapt to the question under processing.

{\bf Meta-learning.}
Meta-learning, aka learning-to-learn, aims to make learning a new task more effective based on the inductive biases that are meta-learned in learning similar tasks in the past. 
%
Huang et al. \shortcite{huang2018natural} use MAML to learn a Seq2Seq model to convert questions in WikiSQL into SQL queries.
More closely related to our work, Guo et al. \shortcite{DBLP:conf/acl/GuoTDZY19} propose Sequence-to-Action (S2A) by using MAML to solve CQA problems.
They label all the examples in training set with pseudo-gold annotations, then train an encoder-decoder model to retrieve relevant samples and a Seq2Seq based semantic parser to generate actions based on the annotations.
Unlike S2A, we introduce a Meta-RL approach, which uses RL to train an NPI model without annotating questions in advance.

\section{Conclusion}
%
CQA refers to answering complex natural language questions on a KB. 
%
%
%
In this paper, we propose a meta-learning method to NPI in CQA, which quickly adapts the programmer to unseen questions to tackle the potential distributional bias in questions. 
We take a meta-reinforcement learning approach to effectively adapt the meta-learned programmer to new questions based on the most similar questions retrieved. 
To effectively create the support sets, we propose an unsupervised retriever to find the questions that are structurally and semantically similar to the new questions from the training dataset. 
%
%
%
When evaluated on the large-scale complex question answering dataset, CQA, our proposed approach achieves state-of-the-art performance with overall macro and micro F1 score of 66.25\% and 77.71\%, respectively. 

In the future, we plan to improve MRL-CQA by designing a retriever that could be optimized jointly with the programmer under the meta-learning paradigm, instead of manually pre-defining a static relevance function.
Other potential directions of research could be toward learning to cluster questions into fine-grained groups and assign each group a set of specific initial parameters, making the model finetune the parameters more precisely. 

%



\section*{Acknowledgments}
This work was partially supported by the National Key Research and Development Program of China under grants (2018YFC0830200), the Natural Science Foundation of China grants (U1736204, 61602259), Australian Research Council (DP190100006), the Judicial Big Data Research Centre, School of Law at Southeast University, and the project no. 31511120201 and 31510040201.

\bibliographystyle{acl_natbib}
\bibliography{anthology,emnlp2020}

\end{document}